\documentclass[letterpaper, 10 pt, conference, twoside]{IEEEconf}

\usepackage{amsfonts,amssymb,amsmath,bm, mathabx}
\usepackage[noend]{algpseudocode}
\usepackage{threeparttable}
\usepackage{enumerate}

\usepackage[ruled,vlined,linesnumbered]{algorithm2e}

\usepackage{tikz}
\usetikzlibrary{matrix,positioning}
\usetikzlibrary{arrows.meta}
\usetikzlibrary{shapes.geometric}
\usetikzlibrary{positioning}
\usetikzlibrary{calc}

\usepackage[pdfencoding=auto]{hyperref}
\hypersetup{
    colorlinks=true,
    linkcolor=black,
    filecolor=magenta,      
    urlcolor=cyan,
    pdftitle={Learning Stable Vector Fields on Lie Groups},
    pdfpagemode=FullScreen,
    }
\usepackage[capitalise]{cleveref}

\usepackage{glossaries}

\SetCommentSty{mycommfont}

\newacronym{rbf}{RBF}{Radial Basis Functions}
\newacronym{rmp}{RMP}{Riemannian Motion Policies}
\newacronym{cep}{CEP}{Composable Energy Policies}

\newacronym{bc}{BC}{Behavioural Cloning}
\newacronym{il}{IL}{Imitation Learning}
\newacronym{rl}{RL}{Reinforcement Learning}
\newacronym{irl}{IRL}{Inverse Reinforcement Learning}
\newacronym{lfd}{LfD}{Learning from Demonstration}
\newacronym{em}{EM}{Expectation Maximization}
\newacronym{promp}{ProMP}{Probabilistic Movement Primitives}
\newacronym{dmp}{DMP}{Dynamic Movement Primitives}
\newacronym{seds}{SEDS}{Stable Estimator of Dynamical Systems}
\newacronym{gmr}{GMR}{Gaussian Mixture Regressor}
\newacronym{gpr}{GPR}{Gaussian Process Regressor}
\newacronym{lwr}{LWR}{Locally Weighted Regressor}
\newacronym{kmp}{KMP}{Kernelized Movement Primitives}
\newacronym{clf}{CLF}{Control Lyapunov Function}
\newacronym{wsaqf}{WSAQF}{Weighted Sum of Asymmetric Quadratic Function}
\newacronym{nilc}{NILC}{Neurally Imprinted Lyapunov
Candidate}
\newacronym{clfdm}{CLF-DM}{Control Lyapunov Function-based Dynamic Movements}
\newacronym{iflow}{iFlows}{ImitationFlows}
\newacronym{cnmp}{CNMP}{Conditional Neural Movement Primitives}
\newacronym{tpgmm}{TP-GMM}{Task Parameterized GMM}

\newacronym{mp}{MP}{Movement Primitive}
\newacronym{mpflows}{MPFlows}{Movement Primitive Flows}

\newacronym{gcl}{GCL}{Guided Cost Learning}

\newacronym{mse}{MSE}{Mean Squared Error}
\newacronym{mle}{MLE}{Maximun Likelihood Estimation}
\newacronym{sde}{SDE}{Stochastic Differential Equation}
\newacronym{ode}{ODE}{Ordinary Differential Equation}

\newacronym{probs}{ProbS}{Probabilistic Segmentation}
\newacronym{crf}{CRF}{Conditional Random Fields}
\newacronym{ppca}{PPCA}{Probabilistic Principal Component Analysis}
\newacronym{gmcc}{GMCC}{Generalized Multiple Correlation Coeficcient}
\newacronym{hri}{HRI}{Human-Robot Interaction}
\newacronym{ip}{IP}{Interaction Primitives}
\newacronym{hmm}{HMM}{Hidden Markov Model}
\newacronym{cac}{CAC}{Canonical Correlation Coefficient}
\newacronym{rv}{$R_v$}{$R_v$ Coefficient}
\newacronym{dcor}{dCor}{Distance Correlation}
\newacronym{dtw}{DTW}{Dynamic Time Warping}
\newacronym{edr}{EDR}{Edit Distance With Real Penalty}
\newacronym{twed}{TWED}{Time Warp Edit Distance}
\newacronym{r2}{$R^2$}{Coefficient of Determination}
\newacronym{sqp}{SQP}{Successive Quadratic Programming}
\newacronym{rkhs}{RKHS}{Reproducing Kernel Hilbert Space}
\newacronym{icnn}{ICNN}{Input-Convex Neural Network}

\newacronym{pca}{PCA}{Principal Component Analysis}

\newacronym{maf}{MAF}{Masked Autoregressive Flow}
\newacronym{iaf}{IAF}{Inverse Autoregressive Flow}
\newacronym{node}{N-ODE}{Neural ODE}
\newacronym{nsflow}{NSF}{Neural Spline Flows}
\newacronym{cnf}{CNF}{Conditional Normalizing Flows}
\newacronym{ffjord}{FFJORD}{Free-form Jacobian of Reversible Dynamics}
\newacronym{inn}{INN}{Invertible Neural Network}

\newacronym{gan}{GAN}{Generative Adversarial Networks}
\newacronym{vae}{VAE}{Variational Autoencoders}
\newacronym{nf}{NF}{Normalizing Flows}

\newacronym{pdf}{PDF}{Probability Density Function}

\newacronym{svf}{SVF}{Stable Vector Field}

\def\1{\bm{1}}

\def\RR{\mathbb{R}}

\def\se{\mathfrak{se}}

\def\d{\text{d}}

\def\vzero{{\bm{0}}}

\def\vtheta{{\bm{\theta}}}

\def\vf{{\bm{f}}}
\def\vg{{\bm{g}}}
\def\vh{{\bm{h}}}

\def\vk{{\bm{k}}}

\def\vp{{\bm{p}}}
\def\vq{{\bm{q}}}

\def\vv{{\bm{v}}}

\def\vx{{\bm{x}}}
\def\vy{{\bm{y}}}
\def\vh{{\bm{h}}}

\def\vpsi{{\boldsymbol{\psi}}}

\def\vtheta{{\boldsymbol{\theta}}}

\def\mA{{\bm{A}}}

\def\mI{{\bm{I}}}
\def\mJ{{\bm{J}}}

\def\mR{{\bm{R}}}

\def\mX{{\bm{X}}}
\def\mY{{\bm{Y}}}

\DeclareMathAlphabet{\mathsfit}{\encodingdefault}{\sfdefault}{m}{sl}
\SetMathAlphabet{\mathsfit}{bold}{\encodingdefault}{\sfdefault}{bx}{n}

\def\gD{{\mathcal{D}}}

\def\gL{{\mathcal{L}}}
\def\gM{{\mathcal{M}}}
\def\gN{{\mathcal{N}}}

\def\gQ{{\mathcal{Q}}}

\def\gS{{\mathcal{S}}}

\makeatletter
\newcommand{\StatexIndent}[1][3]{%
  \setlength\@tempdima{\algorithmicindent}%
  \Statex\hskip\dimexpr#1\@tempdima\relax}
\makeatother

\definecolor{redi}{RGB}{255,105,97}
\definecolor{redii}{RGB}{137,0,3}
\definecolor{yellowi}{RGB}{255,251,0}
\definecolor{bluei}{RGB}{0,150,255}
\definecolor{blueii}{RGB}{135,247,210}
\definecolor{blueiii}{RGB}{91,205,250}
\definecolor{blueiv}{RGB}{115,244,253}
\definecolor{bluev}{RGB}{1,58,215}
\definecolor{orangei}{RGB}{240,143,50}
\definecolor{yellowii}{RGB}{222,247,100}
\definecolor{greeni}{RGB}{166,247,166}

\begin{document}

\title{Learning Stable Vector Fields on Lie Groups}

\author{Julen Urain$^{1}$,  Davide Tateo$^{1}$, and Jan Peters$^{1,2,3,4}$\\
\thanks{Manuscript received: August, 9, 2022; Revised September, 12, 2022; Accepted September, 13, 2022.}
\thanks{$^{1}$ Technische Universität Darmstadt (Germany), 
$^2$ German Research Center for AI (DFKI), $^3$ Hessian.AI, $^4$ Centre for Cognitive Science {\tt\footnotesize\{julen.urain\_de\_jesus, davide.tateo, jan.peters\} @tu-darmstadt.de}}
}


\maketitle


\begin{abstract}
Learning robot motions from demonstration requires models able to specify vector fields for the full robot pose when the task is defined in operational space. Recent advances in reactive motion generation have shown that learning adaptive, reactive, smooth, and stable vector fields is possible. However, these approaches define vector fields on a flat Euclidean manifold, while representing vector fields for orientations requires modeling the dynamics in non-Euclidean manifolds, such as Lie Groups.
In this paper, we present a novel vector field model that can guarantee most of the properties of previous approaches i.e., stability, smoothness, and reactivity beyond the Euclidean space. In the experimental evaluation, we show the performance of our proposed vector field model to learn stable vector fields for full robot poses as SE(2) and SE(3) in both simulated and real robotics tasks. Videos and code are available at:~\url{https://sites.google.com/view/svf-on-lie-groups/}
\end{abstract}


\section{Introduction}
\label{sec:intro}

Data-driven motion generation methods such as \gls{il}~\cite{schaal1997learning, abbeel2004apprenticeship} bring the promise of teaching our robots the desired behavior from a set of demonstrations without further programming of the robot skill. Similar to the CNN networks in computer vision, choosing a good representation of the motion generator might help in the quality of the robot's performance, when learning a policy directly from data. During the last two decades, there has been vast research on learning policy architectures~\cite{schaal2006dynamic, ijspeert2008central,  khansari2011learning, paraschos2013probabilistic, calinon2016tutorial} that guarantee a set of desirable inductive biases. Popularized as \gls{mp}, the community explored a wide set of policy architectures with inductive biases such as Smoothness~\cite{paraschos2013probabilistic}, Stability~\cite{schaal2006dynamic, khansari2011learning} or cyclic performance~\cite{kulak2020fourier, ijspeert2008central}. 

\textbf{Learning Movement Primitives for orientations requires additional insights in the architecture of the model}. There exist multiple representation forms for the orientation, such as Euler angles, rotation matrices, or quaternions. Euler angles have an intuitive representation, but the representation is not unique and might get stuck in singularities (i.e. gimbal lock). These properties make Euler angles undesirable for reactive motion generation~\cite{Yuan1988closed}. Instead of Euler angles, rotation matrices and quaternions are preferred representations for reactive motion generation. Nevertheless, they require special treatment, given they are not defined in the Euclidean space. Rotation matrices are represented by the special orthogonal group, SO(3), while quaternions are represented in the 3-sphere, $\gS^3$. Thus, in the context of modeling orientation \gls{mp}, there has been wide research integrating manifold constraints and \gls{mp}. In \cite{pastor_online_2011, koutras2020correct, ude2014orientation}, \gls{dmp}~\cite{schaal2006dynamic} were adapted to learn orientation \gls{dmp}, by representing \gls{dmp} for quaternions~\cite{pastor_online_2011, koutras2020correct, ude2014orientation} or rotation matrices~\cite{ude2014orientation}.  More recently, orientation \gls{mp} have been also considered to adapt \gls{kmp}~\cite{huang2019kernelized, huang2020toward},  \gls{tpgmm}~\cite{calinon2016tutorial, zeestraten2017approach} and \gls{promp}~\cite{paraschos2013probabilistic,rozo2022orientation}.
Nevertheless, most of the \gls{mp} are rather phase dependant or lack stability guarantees.

\begin{figure}[t]
	\centering
	\begin{minipage}{.5\textwidth}
		\centering
		\includegraphics[width=.99\textwidth]{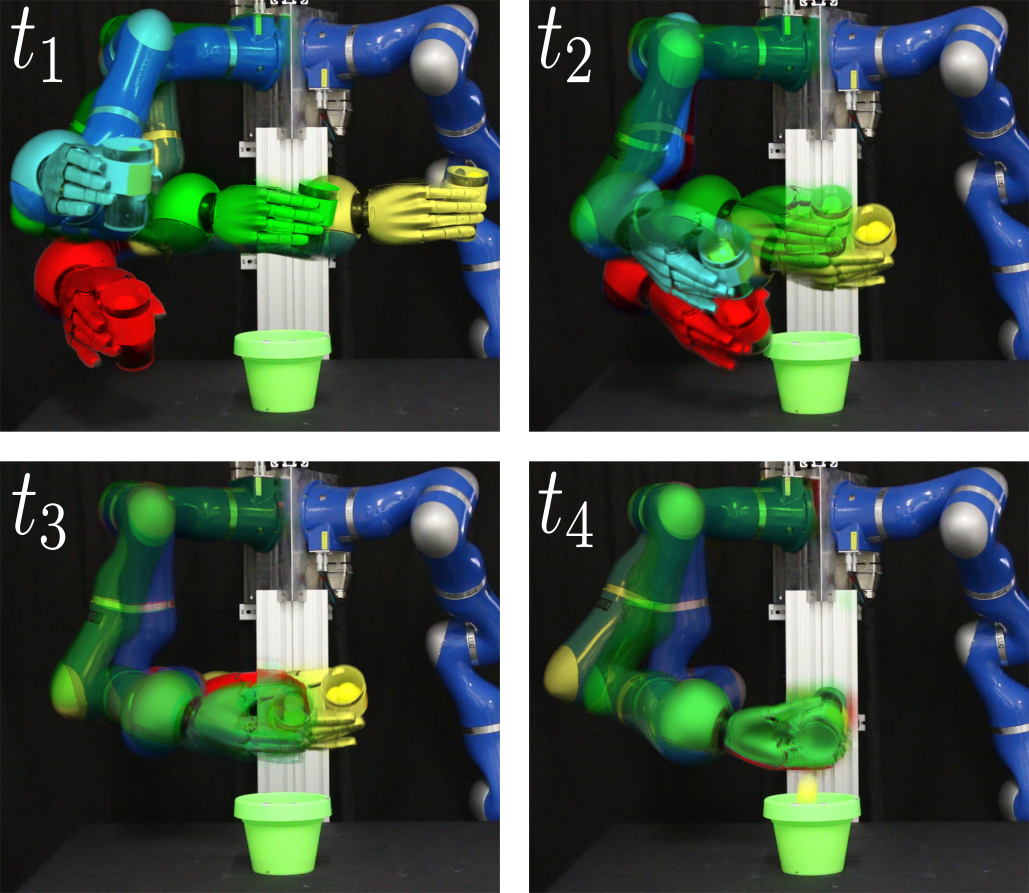}
	\end{minipage}
	\caption{Robot pouring trajectories generated by $SE(3)$-stable vector fields. Each color represents a trajectory starting from a different initial configuration. Given the stability properties, all the trajectories end up with the same orientation and position on the end effector.}
	\label{fig:darias_main}
	\vspace{-.4cm}
\end{figure}  
We propose to learn \gls{svf}~\cite{khansari2011learning, neumann2015learning, urain2020imitation} on position and orientations. \gls{svf} are a family of dynamic systems that are autonomous (i.e. don't have phase dependency, but only depend on the current state) and are inherently stable in terms of Lyapunov. In contrast with \gls{dmp}~\cite{pastor_online_2011} or \gls{kmp}~\cite{huang2020toward}, \gls{svf} are \textbf{inherently reactive to disturbances} without the requirement of any phase adaptation. In contrast with \gls{tpgmm}~\cite{zeestraten2017approach}, \gls{svf} are \textbf{inherently stable}, generating stable motions beyond the expert demonstrations. These properties makes \gls{svf} ideal for human-robot interaction or to combine them with other vector fields as in \gls{rmp}~\cite{ratliff2018riemannian}.

The contribution of this paper are: (1) We introduce a novel learnable \gls{svf} function that can generate stable motions on Lie Groups. Our proposed function generalizes Euclidean space \textsl{diffeomorphism-based \gls{svf}}~\cite{neumann2015learning,urain2020imitation, rana2020euclideanizing} to arbitrary smooth manifolds such as Lie Groups. (2) To learn these \gls{svf}, we propose a neural network architecture that represents diffeomorphic functions in robotic-relevant Lie Groups such as SE(2) and SE(3). (3) Finally, we compare the performance of our proposed model w.r.t. learning the vector fields for Euler angles and learning the vector fields in the configuration space of the robot.

\subsection{Background}
\label{sec:smooth_manifolds}

A n-\textsl{manifold} $\gM$ is called \textsl{smooth} if it is locally diffeomorphic to an Euclidean space $\RR^n$~\cite{lee2006smooth}. For each point $\vx \in \gM$, there exist a coordinate chart $(U,\psi)$, were $U$ is an open subset in the manifold, $U \subseteq \gM$ and $\psi: U \xrightarrow{} \hat{U}$, is a diffeomorphism from the subset $U$ to a subset in the Euclidean space $\hat{U} \subseteq \RR^n$. This chart allows us to represent a section of the manifold $\gM$ in a Euclidean space and do calculus.

\begin{figure}[t]
    \centering
    \includegraphics[width=.45\textwidth]{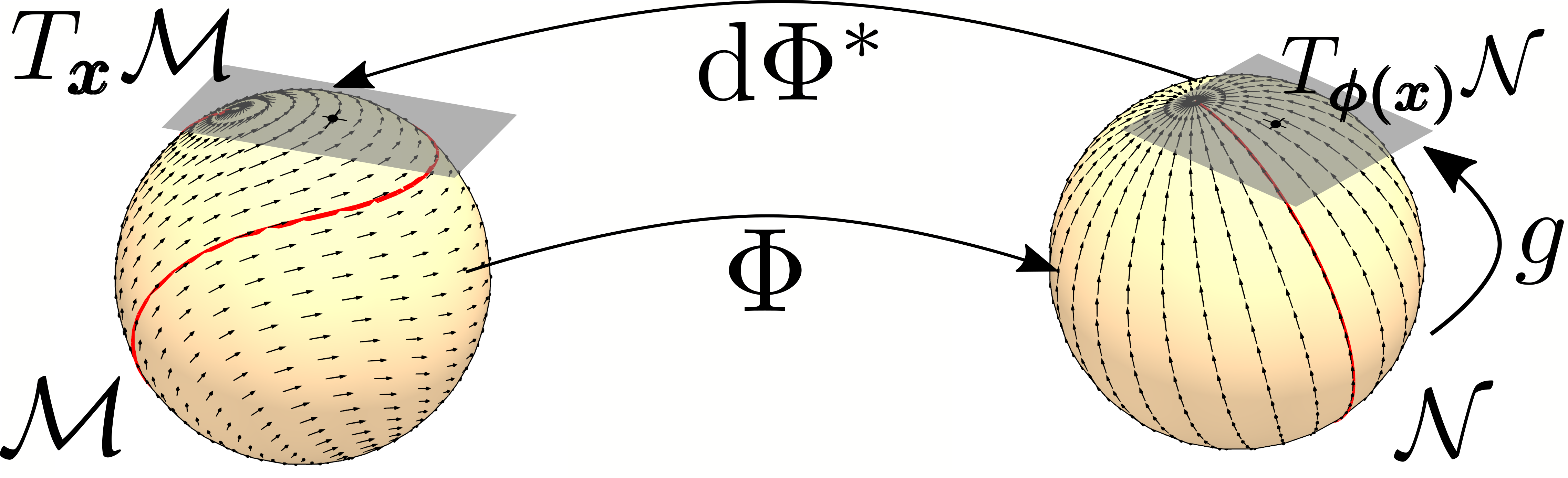}
    \caption{In our work, we compute the vector field in $\gM$ by pulling back the vector field from the latent manifold $\gN$. Given a point $\vx \in \gM$, we first map it to the latent manifold $\vy = \Phi(\vx)$ with $\vy \in \gN$. Then, we compute the vector in the latent manifold. Given a vector field $\vg:\gN \xrightarrow{} T \gN$, we compute $\dot{\vy}\in T_{\Phi(\vx)}\gN$. Finally, we apply the pullback linear operator to compute $\dot{\vx} = \d \Phi_{\vx}^*(\dot{\vy})$ in the tangent space of $\gM$, $\dot{\vx} \in T_{\vx} \gM$. As we can observe, the diffeomorphic function $\Phi$ will deform the space and a trajectory (red line) or a vector field in the manifold $\gN$ will be deformed in $\gM$.}
    \label{fig:main_fig}
\end{figure}

For any point in the manifold, $\vx \in \gM$, we can attach a \textsl{tangent space}, $T_{\vx} \gM$ that contains all the possible vectors that are tangential at $\vx$. Intuitively, for any possible curve in $\gM$ passing through $\vx$, the velocity vector of the curve at $\vx$ will belong to the tangent space, $\vv \in T_{\vx} \gM$. Thus, a \textsl{vector field} in the manifold $\gM$ is a function that maps any point in the manifold to a vector in the tangent space\footnote{The precise term for $T\gM$ is tangent bundle. The tangent bundle is the disjoint set of all tangent spaces. The tangent space is defined at a certain point $\vx$, $T_{\vx}\gM$. For simplicity,  with a slight terminology abuse, in this paper, we use the term tangent space.}, $\vg:\gM \xrightarrow{} T\gM$. The \textsl{LogMap} is the map that moves a point in the manifold $\gM$ to the tangent space, and the \textsl{ExpMap} is the map that moves a point from the tangent space to the manifold.

A map $\Phi: \gM \xrightarrow{} \gN$ between smooth manifolds induces a linear map between their corresponding tangent spaces. For any point $\vx \in \gM$,  the differential of $\Phi$ at $\vx$ is a linear map, $\d \Phi_{\vx}: T_{\vx}\gM,  \xrightarrow{} T_{\Phi(\vx)}\gN$, from the tangent space at $\vx \in \gM$ to the tangent space at $\Phi(\vx) \in \gN$ (Fig.~\ref{fig:main_fig}). 
The differential, $\d \Phi_{\vx}$, is used to map vectors between tangent spaces.  The \textsl{pullback} operator is the linear operation $\d \Phi_{\vx}^{*}: T_{\Phi(\vx)}\gN  \xrightarrow{} T_{\vx} \gM$ that maps a vector from $T_{\Phi(\vx)} \gN$ to $T_{\vx} \gM$.

\subsection{Related Work}
\paragraph{Stable Vector Fields}
\gls{svf} models are powerful motion generators in robotics given they are robust to perturbations and generalize the motion generation beyond the demonstrated trajectories. After the seminal work by Khansari et al.~\cite{khansari2011learning}, several works~\cite{neumann2013neural, perrin2016fast, umlauft2017learning, urain2020imitation, rana2020euclideanizing} have proposed novel \gls{svf} models covering a wider family of solutions. Our work is particularly close to diffeomorphism based \gls{svf} models~\cite{neumann2013neural, perrin2016fast, urain2020imitation, rana2020euclideanizing}. Specifically, in this paper, we extend the class of solutions to non-Euclidean manifolds, such as Lie groups.

\paragraph{Invertible Neural Networks (INN) in Smooth Manifolds} 
\gls{inn} are a family of neural networks that guarantee to represent bijective functions. The study of modeling \gls{inn} for smooth manifolds has been mainly developed for density estimation. A set of previous works ~\cite{rezende2020normalizing, mathieu2020riemannian, gemici2016normalizing} have proposed \gls{inn} for specific manifolds, such as Tori or Sphere manifolds. A more recent work~\cite{lou2020neural} proposes a manifold agnostic approach, on which Neural ODE~\cite{chen2018neural, grathwohl2018scalable} are adapted to manifolds. In \cite{falorsi2019reparameterizing}, \gls{inn} are proposed for Lie Groups. Similar to our work, they also exploit the Lie algebra to learn expressive diffeomorphisms, but the proposed model is limited to density estimation.

\section{Problem Statement}
\label{sec:problem_statement}
We aim to solve the problem of modeling \gls{svf} on Lie Groups. In particular, we model our \gls{svf} by diffeomorphisms. Diffeomorphism-based \gls{svf} represent the vector field in the observation space as the deformed vector field of a certain latent space~\cite{neumann2015learning, perrin2016fast, urain2020imitation, rana2020euclideanizing}. 
These models assume there exist a \textbf{stable vector field in a latent space} $\vg:\gN \xrightarrow{} T\gN$. Then, given a \textbf{parameterized diffeomorphic mapping} $\Phi$, that maps any point in observation space $\gM$ to the latent space $\gN$, $\Phi: \gM \xrightarrow{} \gN$, we can represent the dynamics in the observation space
\begin{align}
        \dot{\vx} = \d \Phi_{\vx}^{*} \circ \vg \circ \Phi(\vx),
        \label{eq:dynamic_proposed}
\end{align}
in terms of the latent dynamics $\vg$ and the diffeomorphism $\Phi$. $\d \Phi_{\vx}^{*}$ is the \textbf{pullback operator} that maps a velocity vector from the latent space to the observation space. Intuitively, as shown in \cref{fig:main_fig}, the diffeomorphic function $\Phi$ deforms the space changing the direction of the vector field in the observation space. The stability guarantees of diffeomorphism-based \gls{svf} have been previously proven in terms of Lyapunov~\cite{urain2020imitation, neumann2015learning}.

Previous \textsl{diffeomorphism-based} \gls{svf} are limited to Euclidean spaces, without representing motion policies in the orientation. Euclidean \gls{svf} assumes (i) that $\Phi:\RR^n \xrightarrow{} \RR^n$ defines a bijective mapping between Euclidean spaces, (ii) in Euclidean spaces, the tangent space and the manifold are in the same space, and then, the latent dynamics are $\vg:\RR^n \xrightarrow{} \RR^n$ and, (iii) given $\Phi$ defines a mapping between Euclidean spaces, the pullback operator is represented by the Jacobian pseudoinverse of $\Phi$, $\d \Phi_{\vx}^{*} = \mJ^{\dagger}_{\Phi}$.

In our work, given we are required to model the \gls{svf} on Lie Groups, we need to (i) model a $\Phi$ function that is bijective between Lie Groups, (ii) investigate how to model stable latent dynamics for Lie Groups and (iii) investigate how to model the pullback operator given the diffeomorphism $\Phi$.

\section{Stable Vector Fields on Lie Groups}
\label{sec:svf}
As introduced in \cref{sec:problem_statement}, modelling \textsl{diffeomorphism-based} \gls{svf} on Lie Groups requires additional insights in the modelling of the three main elements $\Phi$, $\vg$ and $\d \Phi^*$. In the following, we introduce our proposed models to represent each of these elements and we add a  control block diagram on \cref{fig:architecture} to provide intuition on how to use the proposed \gls{svf} in practice.

\subsection{Diffeomorphic Mapping \texorpdfstring{$\Phi$}{Phi}}
\label{sec:diff_model}

We introduce our proposed function to learn diffeomorphisms between Lie Groups, $\Phi: \gM \xrightarrow{} \gN$. Both $\gM$ and $\gN$ are manifolds for the same Lie group, with $\gM$ representing the Lie group in the observation space and $\gN$, the Lie group in the latent space. A simple example of $\Phi$ is given by the rotation function. Given $\mX \in \gM = SO(3)$ and $\mY \in \gN=SO(3)$, the rotation function $\mY = \Phi(\mX) = \mR \mX$, applies a linear diffeomorphic mapping between $\gM$ and $\gN$. 

Nevertheless, representing nonlinear diffeomorphic mappings for Lie groups is challenging.
In our work, we propose to exploit the tangent space to learn these mappings. In contrast with the manifold, the tangent space is a Euclidean space, making it easier to model nonlinear diffeomorphic functions. 

\begin{figure}[t]
	\centering
	\includegraphics[width=.49\textwidth]{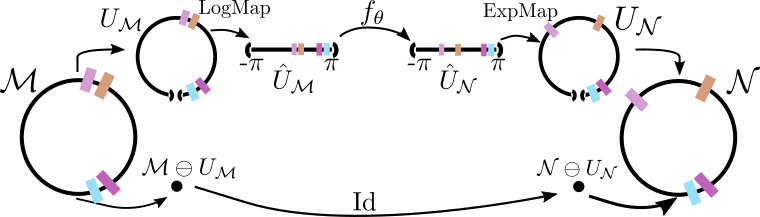}
	\caption{A visual representation of the $\Phi$ function for 1-sphere ($\gS^1$). The points in $\gS^1$ are split into two groups. For the points in $U_{\gM}$, the diffeomorphism is composed by first, mapping the points to the first-cover $\hat{U}_{\gM}$ by the LogMap, then applying a bounded Euclidean diffeomorphism between $\hat{U}_{\gM}$ and $\hat{U}_{\gN}$ and mapping the points back to the manifold, by the ExpMap. For the points not belonging to $U_{\gM}$, we simply apply the identity map. If $f_{\vtheta}$ is the identity map close to the boundaries $-\pi$ and $\pi$; the map is diffeomorphic for the whole $\gS^1$. We add a few markers to represent the space deformation along the mappings.}
	\label{fig:so2_visualization_map}
\end{figure}

\begin{figure*}[t]
    \centering
    \includegraphics[width=.99\textwidth]{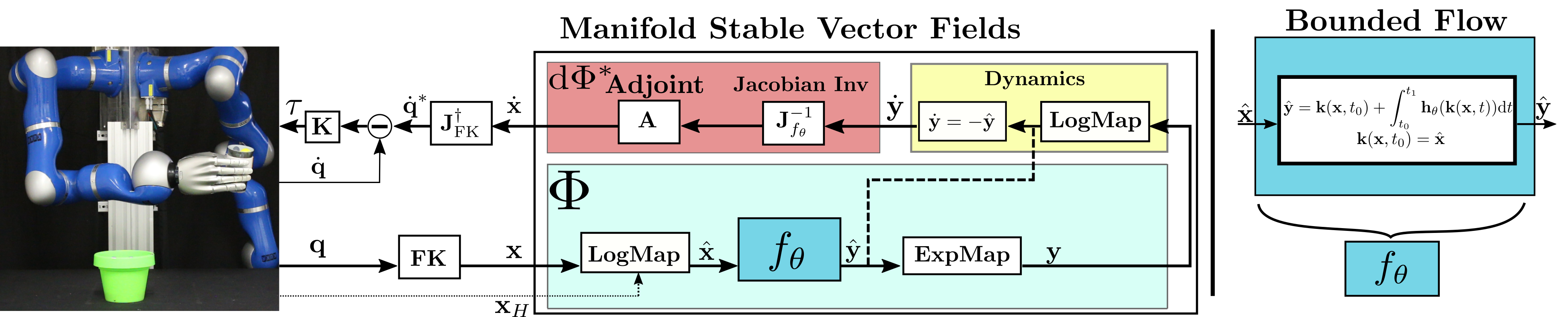}
    \caption{Left: Manifold stable vector fields block diagram. Right: Proposed architecture for our diffeomorphic function $\vf_{\vtheta}$. As shown in \eqref{eq:dynamic_proposed}, our manifold \gls{svf} is composed of three elements: a diffeomorphism $\Phi$ (light blue box)(\textit{for simplicity, we only visualize the part related with the set $U_{\gM}$}), the latent dynamics $g$ (yellow box) and, the pullback operator $\d\Phi^*$ (red box). The diffeomorphism $\Phi$ is composed of three elements: the LogMap, a bounded diffeomorphism between first covers $\vf_{\vtheta}$ (blue) and, the ExpMap. The pullback operator $\d \Phi^*$ has two elements: the Jacobian inverse, computed for the diffeomorphism $\vf_{\vtheta}$, and the Adjoint operator. Additionally, to control a robot, we first map the current joint configuration $\vq$ to $\vx \in SE(3)$ by Forward Kinematics. And once $\dot{\vx} \in \se(3)$ is computed, we map it back to the configuration space by $\mJ^{\dagger}_{\textrm{FK}}$. Then, we apply a velocity controller in the configuration space. The dashed line from the output of $\vf_{\vtheta}$ and the dynamics input represents a shortcut we consider in practice as long as the latent ExpMap and LogMap are computed in the same origin frame.}
    \label{fig:architecture}
    \vspace{-.5cm}
\end{figure*}

The topology of the Lie groups and their Lie algebra are not the same. Then, it is impossible to define a single diffeomorphic function $\Phi$ that maps all the points in the group to the Lie algebra. To make proper use of the Lie algebra and still guarantee the diffeomorphism for the whole Lie group, we propose to model the diffeomorphism by parts. We visualize an example of the proposed function in \cref{fig:so2_visualization_map}. The points in the Lie Group are split into two sets. We consider a coordinate chart $U_{\gM} \subseteq \gM$ that defines a set of almost all the points in the Lie group. Then, we group all the points not belonging to the set $ U_{\gM}$ in a different set, $\vx \in \gM \ominus U_{\gM}$. For example, in the example on \cref{fig:so2_visualization_map}, we group all the points except the antipodal point in $U_{\gM}$ and put the antipodal point in the set $\gM \ominus U_{\gM}$.
The points in the set $U_{\gM}$ are mapped to a set in the latent manifold, $U_{\gN} \subseteq \gN$. The points in the set $\gM \ominus U_{\gM}$ are mapped to the latent space set $\gN \ominus U_{\gN}$. Given that $\gM$ and $\gN$ are represented in the same Lie Group, the sets in the observation space and the latent space are also the same.
\begin{align}
\label{eq:diff_parts}
       \Phi(\vx) = \begin{cases} \textrm{ExpMap}\circ \vf_{\vtheta} \circ \textrm{LogMap} (\vx) &\mbox{if } \vx \in U_{\gM} \\
\vx & \mbox{if }\vx \in \gM \ominus U_{\gM}. \end{cases}
\end{align}
For any element in the coordinate chart $\vx \in U_{\gM}$, we define the map from $U_{\gM}$ to $U_{\gN}$, through the tangent space, $\Phi:\textrm{ExpMap}\circ \vf_{\vtheta} \circ \textrm{LogMap}$. The function first maps a point in the Lie group to the Lie algebra by the $\textrm{LogMap}$. For any point $\vx \in U_{\gM}$, it will map to a point in a subset of the tangent space, $\hat{\vx} \in \hat{U}_{\gM} \subseteq  T_{\vx_H} \gM $. We call \textbf{first cover of the tangent space} to $\hat{U}_{\gM}$. The map between $U_{\gM}$ and $\hat{U}_{\gM}$ is guaranteed to be diffeomorphic given the LogMap properties~\cite{lee2006smooth}. Then,
we apply a Euclidean diffeomorphism  $\vf_{\vtheta}$ between the first covers of the observation space $\hat{U}_{\gM}$ and the first covers of the latent space, $\hat{U}_{\gN}$. We introduce our proposed $\vf_{\vtheta}$ in \cref{sec:invertible_network}. Finally, we can map the points $\hat{\vy} \in \hat{U}_{\gN}$ back to $\vy \in U_{\gN} \subseteq \gN$ by the ExpMap and represent it in the Lie Group. Given the three steps are diffeomorphic, we can guarantee that $\Phi$ applies a diffeomorphism between $U_{\gM}$ and $U_{\gN}$. For the points not belonging to the set $U_{\gM}$, we apply the identity map. The identity map is also diffeomorphic.

Even if each part in \cref{eq:diff_parts} is diffeomorphic in itself, to guarantee the function $\Phi$ is diffeomorphic in the whole Lie group, we require to guarantee the function is continuous and differentiable in the boundaries between $U_{\gM}$ and $\gM \ominus U_{\gM}$. To do so, we impose structurally $\vf_{\vtheta}$ to become the identity map $\vf_{\vtheta}(\hat{\vx}) = \hat{\vx}$ when approaching to the boundaries of the set $\hat{U}_{\gM}$. Thus,
\begin{align}
    \Phi(\vx) &=  \textrm{ExpMap}\circ \vf_{\vtheta} \circ \textrm{LogMap} (\vx)\nonumber \\ &=  \textrm{ExpMap} \circ \textrm{LogMap} (\vx) = \vx
\end{align}
when $\vx$ is close to the boundaries of $U_{\gM}$.

\subsubsection{An intuitive example for 1-sphere (\texorpdfstring{$S^1$}{S1}) manifold}
The 1-sphere manifold is composed by all the points in a circle of radius $r$, $\gS^1:\{\vx \in \RR^2 \,;\, ||\vx||=r \}$. We visualize this manifold in \cref{fig:so2_visualization_map}.
To model a diffeomorphic transformation between $\gM$ and $\gN$, we propose to split the manifold in two sets: the set $U_{\gM}$ considers all the points in the manifold except the point in the south $\vx_{S} = (0, r)$, $U_{\gM} = \gS^1_{\neq \vx_S}$. Equally, the set in the latent manifold $U_{\gN}$, also consider $U_{\gN} = \gS^1_{\neq \vx_S}$. 
The other set $\gM \ominus U_{\gM} = \{ \vx_{S} \}$ is composed of the point not belonging to $U_{\gM}$.
We can observe that $U_{\gM}$ is diffeomorphic to the open line segment $\hat{U}_{\gM}=(-\pi, \pi)$. We refer to this set as first-cover of the tangent space $\hat{U}_{\gM} = \hat{U}_{\gN} = (-\pi,\pi)$. We can map any point from $U_{\gM}$ to $\hat{U}_{\gM}$ by the LogMap function. Inversely, we can map the points from the open line segment to the set $U_{\gM}$ by the ExpMap function. We remark that points in $U_{\gM}$ are two-dimensional while points in $\hat{U}_{\gM}$ are one-dimensional. 
Once the points are in the $\hat{U}_{\gM}$, we model a bounded diffeomorphic function $\vf_{\vtheta}$ that maps the points in $\hat{U}_{\gM}$ to $\hat{U}_{\gN}$. We present in \cref{sec:invertible_network} how we model this bounded diffeomorphism $\vf_{\vtheta}$. This map can be thought of as a deformation of the line $\hat{U}_{\gM}$, stretching or contracting the line. 
We highlight that while representing directly a diffeomorphism between the open line segments $\hat{U}_{\gM}$ and $\hat{U}_{\gN}$ is easy, representing it between the groups $U_{\gM}$ and $U_{\gN}$ is hard, given that $U_{\gM}$ and $U_{\gN}$ are not Euclidean spaces.

As shown before, to guarantee that $\Phi$ is diffeomorphic for the whole manifold $\gS^1$, we need to guarantee that $\vf_{\vtheta}$ becomes the identity map close to the boundaries of $\hat{U}_{\gM}$. For the case of $\gS^1$, the function $\vf_{\vtheta}$ should approximate the identity map the closer the points are to $-\pi$ and $\pi$. Intuitively, the function $\vf_{\vtheta}$ represents a space deformation in $(-\pi, \pi)$ that becomes the identity close to the boundaries $-\pi$ or $\pi$. We illustrate this diffeomorphic map in \cref{fig:so2_visualization_map}.

\subsection{Latent Stable Dynamics \texorpdfstring{$g$}{g}}
For a given manifold $\gN$, the vectors are represented in the tangent space of the manifold, $T\gN$. 
Thus, a dynamic system in a manifold is a function that for any point in the manifold outputs a vector in the tangent space, $\vg: \gN \xrightarrow{} T \gN$.
Similarly to the transformation map $\Phi$, we propose to model the dynamics by parts
\begin{align}
    \label{eq:latent_dynamics}
    \dot{\vy} = g(\vy) = \begin{cases}  -\textrm{LogMap}_{\vy_H} (\vy) &\mbox{if } \vy \in U_{\gN} \\
     \vzero & \mbox{if }\vy \in \gN \ominus U_{\gN} \end{cases}.
\end{align}
For any element in $U_{\gN}$, we first map the point to the tangent space centered at $\vy_H$ and then, compute the velocity vector as $\dot{\vy} = \vg(\hat{\vy}) = - \hat{\vy}$.
These dynamics will induce a stable dynamic system in the manifold $U_{\gN}$, with a sink in $\vy_H$. For any point out of the set $U_{\gN}$, we set the velocity to zero. This will set an unstable equilibrium point for any point in $\gN \ominus U_{\gN}$. In practice, given the LogMap in our dynamics \cref{eq:latent_dynamics} is the inverse of the ExpMap in $\Phi$, we can directly compute the dynamics using as input the output of $f_{\vtheta}$ without moving to $\gN$~(dashed line in \cref{fig:architecture}).

\subsection{Pullback Operator \texorpdfstring{$\d \Phi^{*}$}{dPhi}}
The pullback operator unrolls all the steps to the latent space, $\gN$, done by the diffeomorphism, $\Phi$, back to the observation manifold, $\gM$. Additionally, given the velocity vector is defined on the tangent space, the unrolling steps are done on the tangent space. The pullback operator for the mapping, $\vf_{\vtheta}$, is the Jacobian $\mJ_{\vf}$. The inverse of the Jacobian, maps the velocity vector from the latent tangent space to the observation tangent space, centered in the origin, $\mJ_{\vf}^{-1} : T_{\vy_H} \gN \xrightarrow{} T_{\vx_H} \gM$. Additionally, we apply a second pullback operator to map the vector from the tangent space in the origin $\vx_H$ to the tangent space in the current pose $\vx$, $\mA: T_{\vx_H} \gM \xrightarrow{} T_{\vx} \gM$. This linear map is known as the adjoint map and it can be understood as a change of reference frame for the velocity vectors. We direct the reader to \cite{sola2018micro} to find more information on how to model it. The whole pullback operator is then, $\d \Phi^* = \mA \circ \mJ_{\vf}^{-1}$.

\begin{algorithm}[t]
    \small
	\SetKwInOut{Input}{Given}
	\Input{$\textrm{mSVF}$: Manifold \gls{svf} function \;
		$\vtheta_0$: initial parameters of the function $\textrm{mSVF}$\;
		$I$: Optimization steps\;
		$\gD:\{\{\vx_{i,t}, \dot{\vx}_{i,t}\}_{t=1}^{T_i}\}_{i=1}^{N}$: $N$ trajectories, of $T_i$ length, with the position in $\vx \in \gM$ and  the velocity vector in $\dot{\vx} \in T_{\vx} \gM$}
	\BlankLine
	\For{$i \leftarrow 0$ \KwTo $I-1$}{
		$\vx_b, \dot{\vx}_b \sim \gD$\tcp*{Sample a batch from dataset}
		$\gL(\vtheta_i) = \frac{1}{B}\sum_{k=0}^B ||\dot{\vx}_k - \textrm{mSVF}(\vx_k;\vtheta_i) ||_2^2$ \;
		$\vtheta_{i+1} \leftarrow \vtheta_{i} + \alpha \nabla_{\vtheta} \gL(\vtheta_i)$\;
		}
	\KwRet{$\vtheta^*$}\;
	
	\caption{Behavioural Cloning for Manifold-\gls{svf} \label{alg:train_m_svf}}
\end{algorithm}

\section{Bounded Flows as transformation \texorpdfstring{$\vf_{\vtheta}$}{f}}
\label{sec:invertible_network}
In \cref{sec:diff_model}, we propose to model the diffeomorphism between two subsets of the manifolds~($U_{\gM}$ and $U_{\gN}$) through the tangent space. To properly model the diffeomorphism, we have introduced a function $\vf_{\vtheta}$ and defined its required properties. The function $\vf_{\vtheta}$ should be a diffeomorphism and should become identity when approximating the boundaries of the tangent space sets $\hat{U}_{\gM}$ and $\hat{U}_{\gN}$. To represent our function $\vf_{\vtheta}$, we build on top of the research on \gls{inn} for Normalizing Flows~\cite{rezende2015variational, chen2018neural}.

We propose to model the function $\vf_{\vtheta}$ by adapting Neural ODEs~\cite{chen2018neural} to our problem. Neural ODEs propose to model the diffeomorphism between two spaces by the flow of a parameterized vector field $\vh_{\vtheta}$. The flow  $\vk(\vx, t):\RR^{n+1} \xrightarrow{}\RR^n$, represents the motion of a point for the time $t$, given the ODE, $\d \vx / \d t = \vh_{\vtheta}(\vx) \equiv \d (\vk(\vx, t)) / \d t = \vh_{\vtheta}(\vk(\vx, t))$
\begin{align}
    \vx_{t_1} = \vk(\vx, t_1) = \vx + \int_{0}^{t_1} \vh_{\vtheta}(\vk(\vx, t)) \d t.
\end{align}
with $t_1$ being a certain time instant and $\vx$ the position of the particle in the instant $t=0$. The flow function represents the position of a particle $\vx$ follows given the vector field $\vh_{\vtheta}$ at the instant $t_1$. 
In Neural ODEs, the function $\vf_{\vtheta}$ is represented by the output of the flow at time $1$
\begin{align}
    \label{eq:neural_ode_solve}
    \vy = \vf_{\theta}(\vx) = \vk(\vx, t=1) = \vx + \int_{0}^{1} \vh_{\vtheta}(\vk(\vx, t)) \d t.
\end{align}
As presented in \cite{mathieu2020riemannian, chen2018neural}, the function is a diffeomorphism, as long as $\vh_{\vtheta}$ is a uniformly Lipschitz continuous vector field~(Picard–Lindelöf theorem). 

Additionally, to compute the pullback operation, we are required to compute the Jacobian matrix of $\vf_{\vtheta}$, $\mJ_{f} = \nabla_\vx \vk(\vx,t_1)$. Given the vector field $\vh_{\vtheta}$, there exists an ODE representing the time evolution of the Jacobian
\begin{align}
    \label{eq:J_neural_ode_solve}
    &\dot{\mJ_{f}}(\vx, t) = \nabla_{\vk}\vh_{\vtheta}(\vk(\vx,t)) \mJ_{f}(\vx,t) \nonumber \\
    & \mJ(\vx,t_0)= \mI.
\end{align}
In practice, we can use an arbitrary ODE solver and find the values for $\mJ(\vx,t_1)$ and $\vk(\vx,t_1)$ solving \eqref{eq:neural_ode_solve} and \eqref{eq:J_neural_ode_solve}. In our case, to guarantee a high control frequency rate, we apply the forward Euler method to solve the ODE and then compute the Jacobian by backward differentiation.
It is important to remark that these dynamics are used to represent the diffeomorphism $\vf_{\vtheta}$ between two spaces and not to represent the desired vector fields.

Relevant consideration for our problem is that the function $\vf_{\vtheta}$ should define a diffeomorphism between two bounded sets $\hat{U}_{\gM}$ and $\hat{U}_{\gN}$ and the transformation should become identity close to the boundaries of these sets. Nevertheless, without any additional considerations on $\vh_{\vtheta}$, the flow could move a point in $\hat{U}_{\gM}$ to any point in $\RR^{n}$, with $n$ the dimension of the Euclidean space in which the set $\hat{U}_{\gN}$ is. To bound the flow between the sets, we impose structurally that the vector field $\vh_{\vtheta}$ vanishes when approaching the boundaries. If the flow dynamics are zero, then, the input and the output are the same and we don't apply space deformation at that point. Given a distance function $\alpha(\vx):\RR^n \xrightarrow{} \RR$ that measures how close we are to the boundaries, we define the vector field as
\begin{align}
    \vh_{\vtheta}(\vx) = \alpha(\vx)\vpsi_{\vtheta}(\vx),
\end{align}
with $\vpsi$ an arbitrarily chosen uniformly Lipschitz continuous parameterized vector field and $\alpha$ the scaling function of the dynamics to satisfy the desired constraints, preventing to move out of the set. $\alpha$ becomes zero close to the boundaries. Then, close to the boundaries, 
\begin{align}
\vy = \vf_{\vtheta}(\vx) = \vk(\vx, t_1) \approx \vk(\vx, t_0) = \vx.
\end{align}
Thus, the function $\vf_{\vtheta}$ is guaranteed to approximate the identity in the boundaries. 

Given the set $\hat{U}_{\gM}$ varies between the manifolds, we consider different distance functions $\alpha$ for each possible manifold. For the case of $SO(2)$, the first covers are $\hat{U}_{\gM}=\hat{U}_{\gN}=(-\pi, \pi)$. To impose identity map in the boundaries, the dynamics are weighted with $\alpha(x) = (\pi - |x|)/\pi$. $\alpha$ is a function that moves from 1 to 0 when we approach the $\pm\pi$ boundaries.

For the case of $\gS^2$, the sets are $\hat{U}_{\gM}=\hat{U}_{\gN}=\{ \vx \in \RR^2 ; ||\vx||<\pi \}$. This set is diffeomorphic to the set in $U_{\gM} = U_{\gN} \subset \gS^2$, which considers all the points in the manifold except the antipodal point. To impose the dynamics to become zero close to the boundaries of the set, the distance function is $\alpha(\vx) = (||\vx||-\pi)/\pi$.

For the case of $SO(3)$, the sets are $\hat{U}_{\gM}=\hat{U}_{\gN}=\{ \vx \in \RR^3 ; ||\vx||<\pi \}$. 
The sets are diffeomorphic to the SO(3) sets $U_{\gM} = U_{\gN} =  SO(3)_{\neq\mR_{\pi}} \subset SO(3)$, that consider all possible rotation matrices except the ones that have a $\pi$ rotation from the origin. The dynamics are weighted by the function $\alpha(\vx) = (||\vx||-\pi)/\pi$.

For the case of the special Euclidean groups SE(2) and SE(3), the orientation-related dimensions maintain the same first covers of the special orthogonal groups. For the position-related dimensions, we bound the first cover to the desired workspace. Given $(\vp, \vtheta) \in \se (3)$, with $\vp$ the position related variables and $\vtheta$, orientation related variables. We consider two scaling functions, one for orientations and one for positions. The orientation scaling function $\alpha_{\text{ori}}(\vtheta)$ is computed given the scaling functions above. The scaling function for the positions  $\alpha_{\text{pos}}(\vp)$ can be used to enforce workspace limits and varies depending on the chosen workspace boundaries. We compute the distance function  by $\alpha(\vp, \vtheta) = \alpha_{\text{pos}}(\vp)\alpha_{\text{ori}}(\vtheta)$.

\begin{figure*}[t]
	\centering
	\begin{minipage}{.24\textwidth}
		\centering
		\includegraphics[width=.99\textwidth]{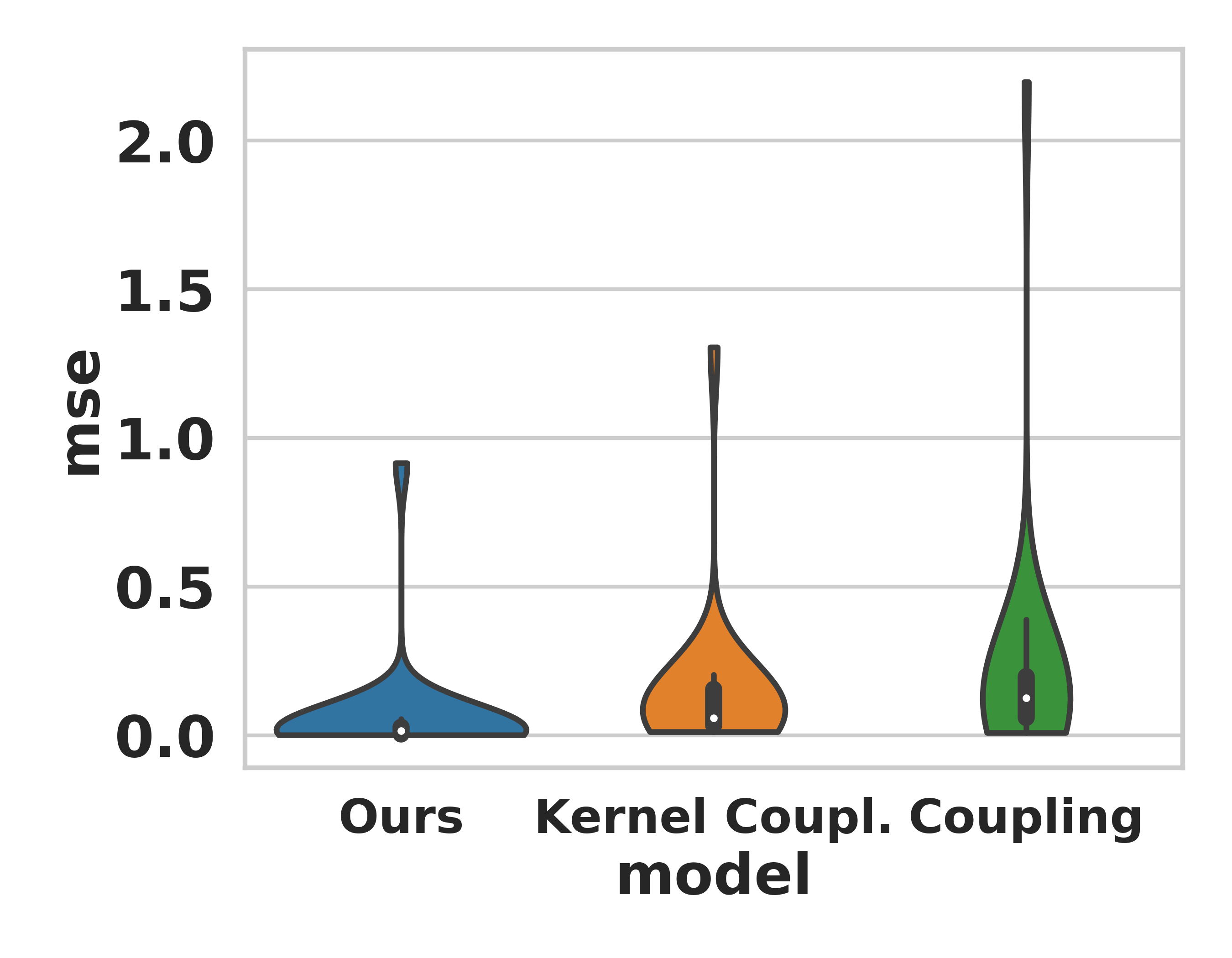}
	\end{minipage}
	\begin{minipage}{.24\textwidth}
		\centering
		\includegraphics[width=.99\textwidth]{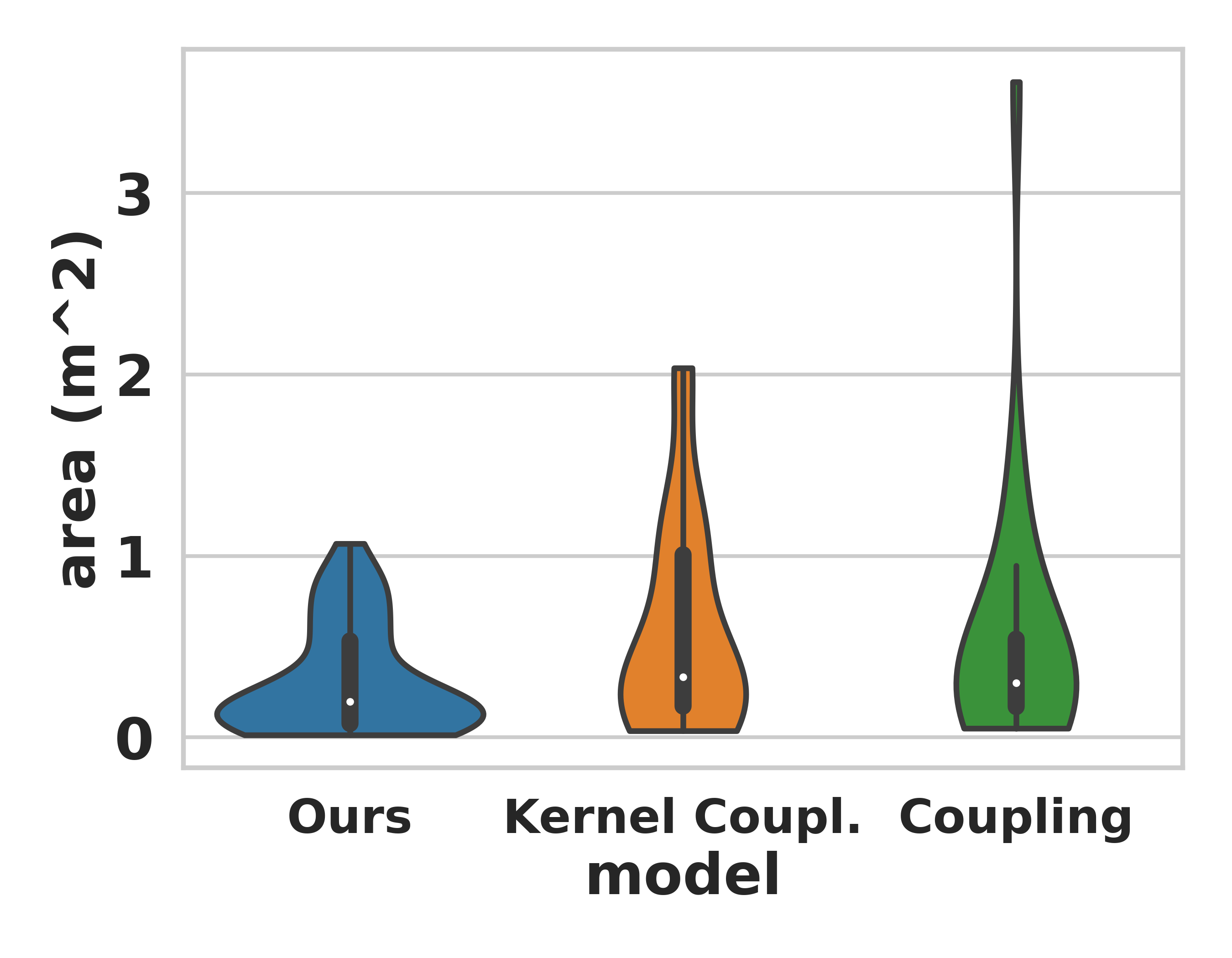}
	\end{minipage}
	\begin{minipage}{.24\textwidth}
		\centering
		\includegraphics[width=.99\textwidth]{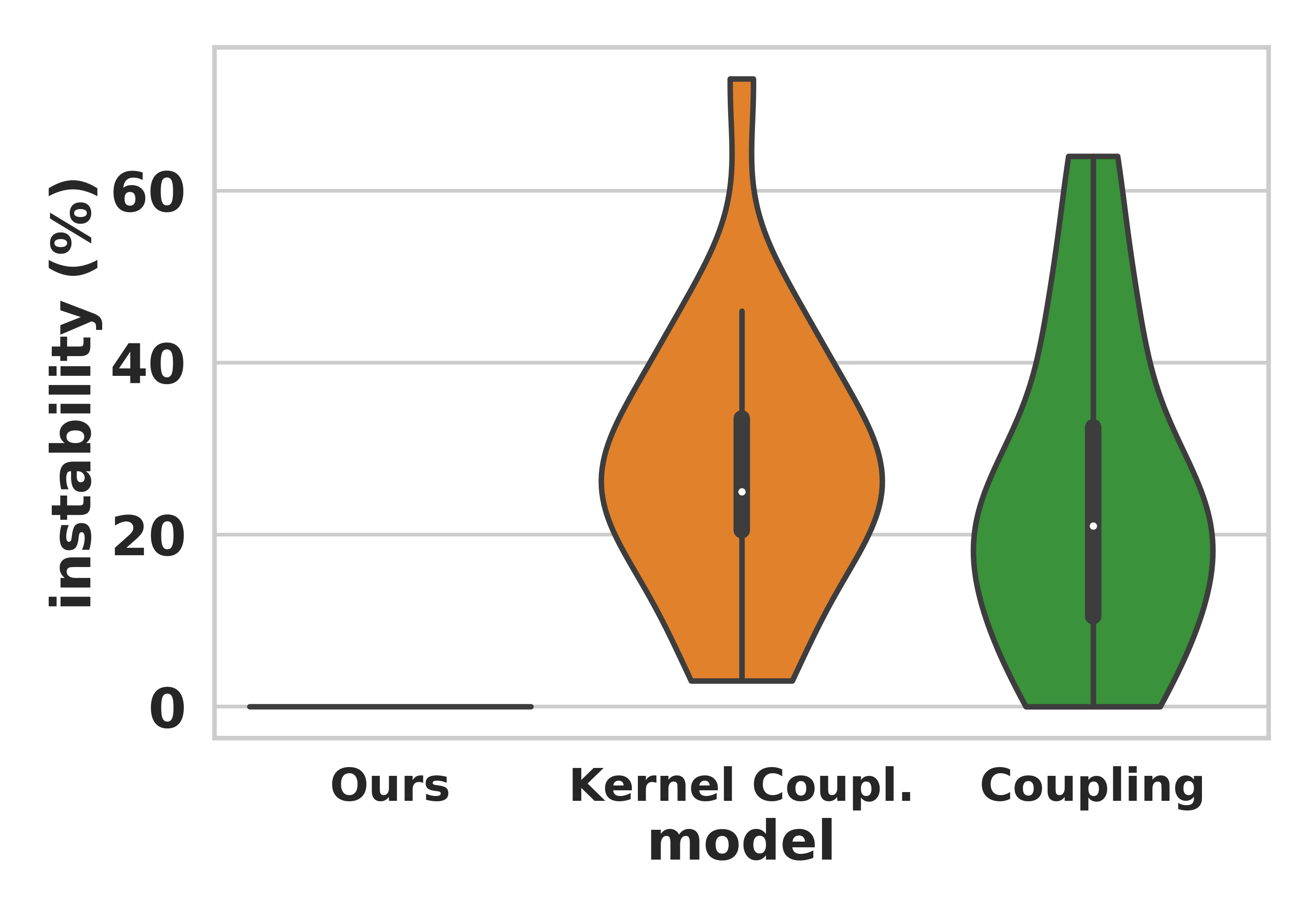}
	\end{minipage}
	\begin{minipage}{.24\textwidth}
		\centering
		\includegraphics[width=3.cm, height=3.cm]{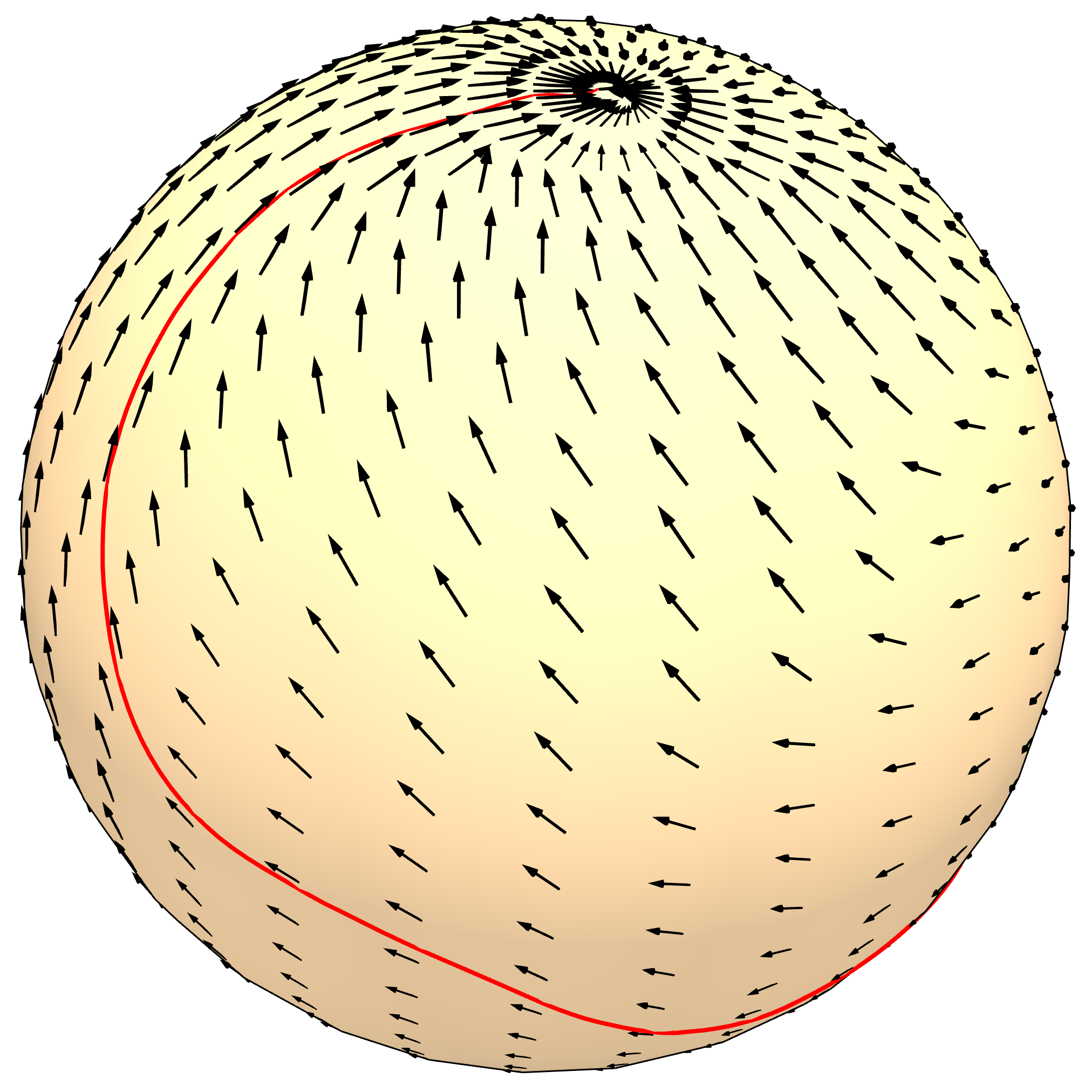}
	\end{minipage}
	\caption{Left: Kernel Coupling, Coupling, and Ours(Smooth Piecewise Linear) Layers compared in terms of \gls{mse}, Area and Instability \%. Kernel Coupling and Coupling Layer apply a diffeomorphism between $\RR^n$ and Ours between the first covers. Right: Example of LASA trajectory and learned vector field.}
	\label{fig:s2_lasa_architecture}
\end{figure*}

\section{Experimental Results}
\label{sec:result}

We present three experiments to evaluate the performance of our approach. In the first experiment, we illustrate, in a $\gS^2$ manifold, the performance of our proposed $\vf_{\vtheta}$ w.r.t. functions that do not take into consideration the manifold and treat is as Euclidean. Even if $S^2$ is not a Lie Group, we can apply the proposed approach also on it and serves as a useful manifold for illustration.

In the second and third experiments, we evaluate the performance of our model in the Lie Groups SE(2) and SE(3), for a 2D peg-in-a-hole task and a pouring task respectively.

\subsection{Network Evaluation in \texorpdfstring{$S^2$}{S2} manifold}
\label{sec:s2_vector_field}

We study the problem of learning stable vector fields in 2-sphere, $S^2$ by behavioral cloning (\cref{alg:train_m_svf}). The objective of this experiment is to evaluate the influence of choosing different \gls{inn} as mapping $\vf_{\vtheta}$.

For evaluation, we consider three models. The three models use our  proposed architecture in \cref{fig:architecture} and vary in the used diffeomorphism $\vf_{\vtheta}$. We consider two models using the \gls{inn} from previous works \cite{rana2020euclideanizing, urain2020imitation} that considers a diffeomorphism in the whole Euclidean space $\vf_{\vtheta}: \RR^n \xrightarrow{} \RR^n$ and our proposed \gls{inn} that learns a diffeomorphism in bounded domains, $\vf_{\vtheta}: \hat{U}_{\gM} \xrightarrow{} \hat{U}_{\gN}$. We modified the LASA dataset~\cite{khansari2011learning} to $\gS^2$ manifolds. We consider 22 different shape trajectories and evaluate the models given three metrics: \gls{mse}, Area, and Instability percentage. For measuring the instability percentage, we initialized a set of points in random positions on $S^2$ and generated a trajectory with the learned vector fields. Then, we measured how many trajectories reach the target position after a certain period. 

\begin{figure}[b]
	\centering
	\begin{minipage}{.4\textwidth}
		\centering
		\includegraphics[width=.99\textwidth]{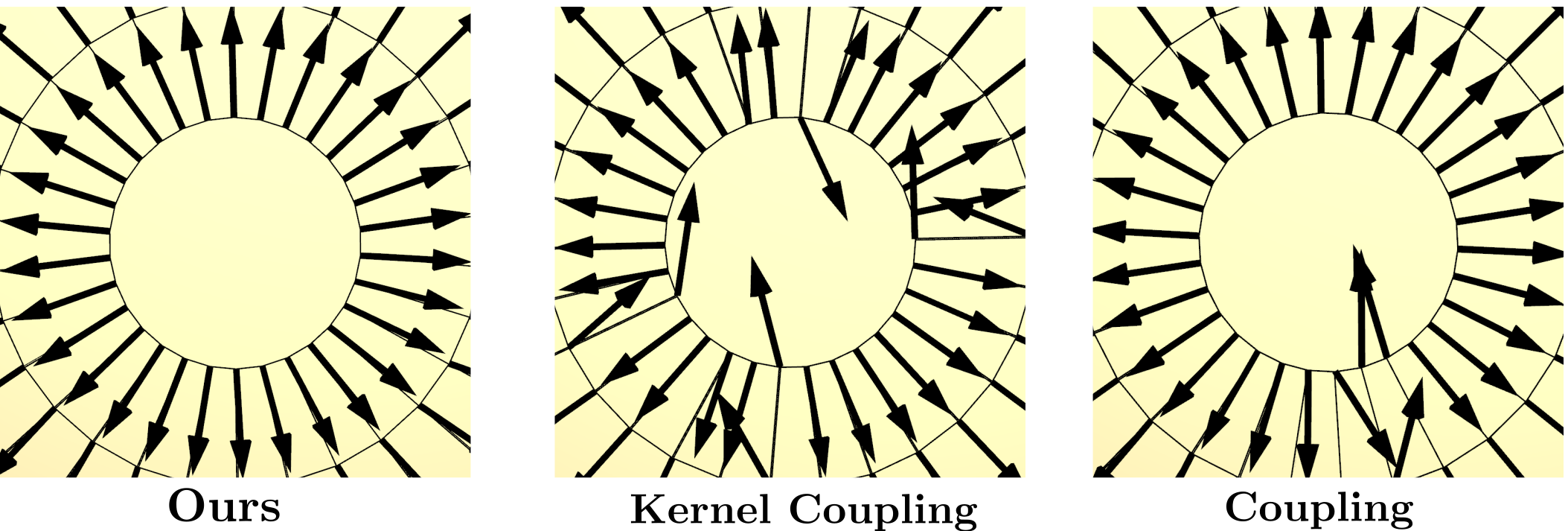}
	\end{minipage}
	\caption{Vector fields in the antipodal point of the Sphere. Our proposed diffeomorphism guarantees a source in the antipodal, while the unbounded \gls{inn} does not.}
	\label{fig:antipodals}
\end{figure}

From \cref{fig:s2_lasa_architecture}, we can observe that the three architectures performed similarly in both \gls{mse} and Area measures and were able to mimic the performance of the demonstrations properly. This indicates that the proposed algorithm can learn vector fields on smooth manifolds. 
Nevertheless, as shown in the Instability $\%$ metric, the performance of the Kernel Coupling Layer~\cite{rana2020euclideanizing} and the Coupling Layer~\cite{urain2020imitation} decay when initializing the trajectories in a random position. Given the Kernel Coupling Layer and the Coupling Layer define a diffeomorphism in the whole Euclidean space, they lack any guarantee of being bijective between $\hat{U}_{\gM}$ and $\hat{U}_{\gN}$. Thus, these approaches lack guarantees about the stability of the vector field in $\hat{U}_{\gM}$. 
We can observe the instability of the vector fields by observing the antipodal point of the sphere, where the boundaries of the first cover $\hat{U}_{\gM}$ are defined. As shown in \cref{fig:antipodals}, while our \gls{inn} can guarantee all the vectors pointing out of the antipodal (a source in the antipodal point), the kernel coupling layer and coupling layer are not able to guarantee stability close to the boundaries generating oscillatory behaviors around the antipodal point. 

\subsection{Evaluation of \texorpdfstring{$SE(2)$}{SE(2)} Stable vector fields in a 2D peg-in-a-hole task}

\begin{figure}[b]
	\centering
	\begin{minipage}{.25\textwidth}
		\centering
		\includegraphics[width=.99\textwidth, height=2.5cm]{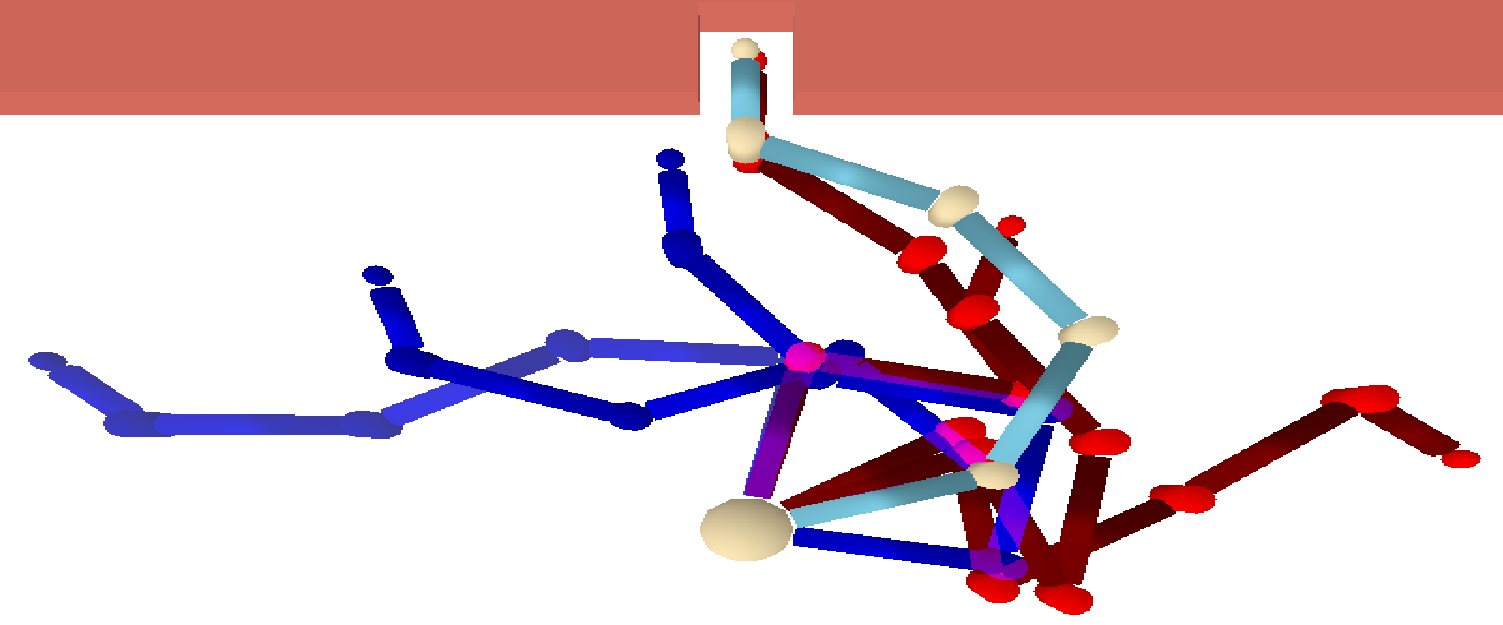}
		\vspace{.7cm}
	\end{minipage}
	\begin{minipage}{.22\textwidth}
		\centering
		\includegraphics[width=.99\textwidth]{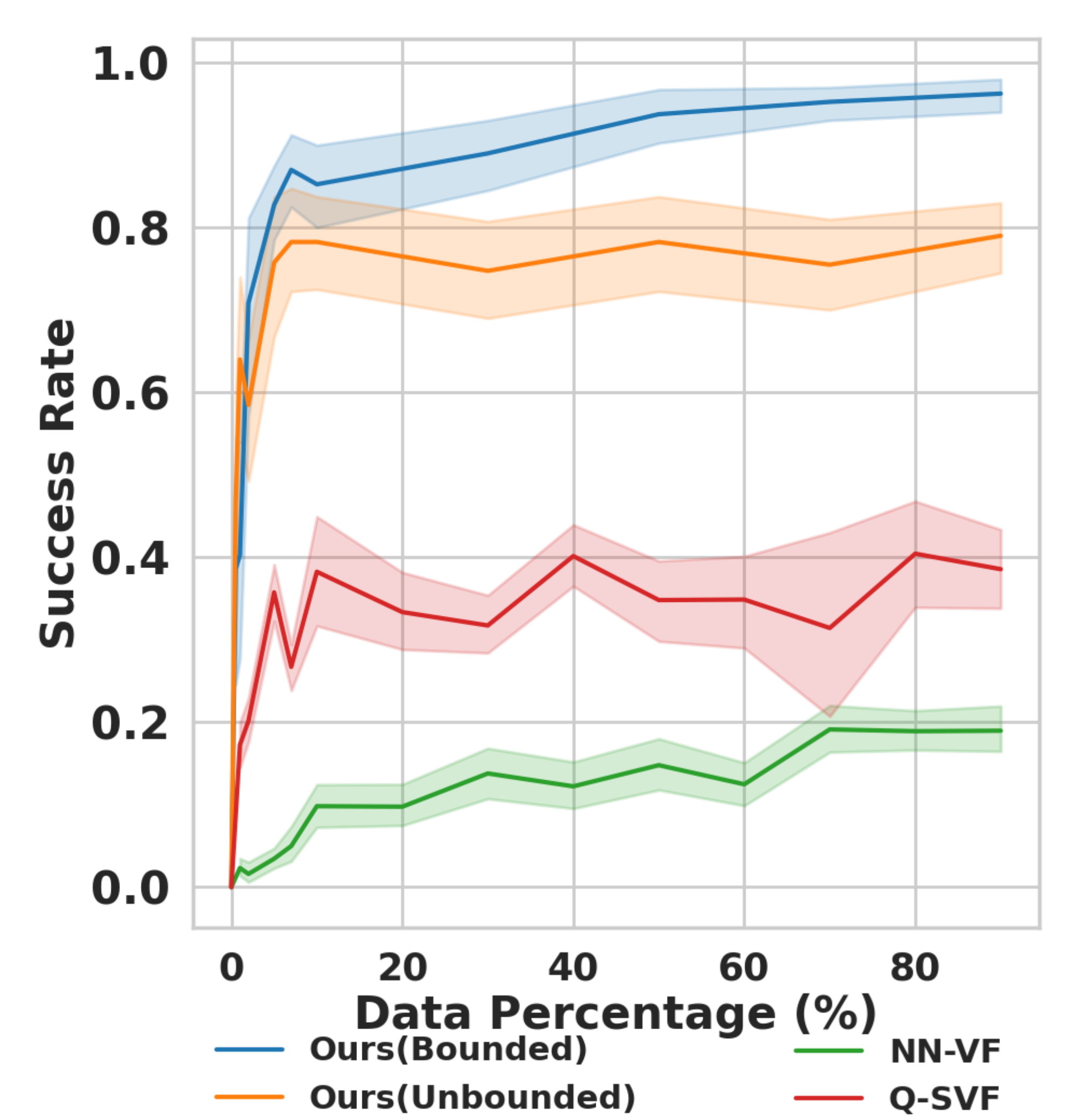}
	\end{minipage}
	\caption{Left: Peg-in-a-hole environment. We show in different colors, generated trajectories from different initial configurations. Right: Success rate Vs. Data percentage. We evaluate the performance of a set of models when trained with different amounts of data.}
	\label{fig:peg_hole_results}
\end{figure}

We consider the environment presented in \cref{fig:peg_hole_results}. The robot is a 5-DOF robot moving in a 2d plane. The goal of the task is to move the end-effector of the robot into the hole while avoiding collisions against the walls. We generated a 1K trajectory demonstration to train our models by applying  RRT-Connect~\cite{kuffner2000rrt} on the environment. We compare the performance of our model w.r.t. three baselines. First, we consider a vector field modeled by a naive fully connected neural network in the tangent space of $SE(2)$. 
Second, we trained a stable vector field in the configuration space, $\gQ$. Third, similarly to the experiment in $\gS^2$, we model a vector field with the architecture in \cref{fig:architecture}, but consider a vanilla \gls{inn} as $\vf_{\vtheta}$ instead of the proposed \gls{inn}.
To evaluate the performances, we initialize the robot in a random configuration and reactively evolve the dynamics. To control the robot, we apply operational space control~\cite{khatib1987unified}. Given the current end-effector pose, $\vx \in SE(2)$, we compute the desired velocity at the end effector $\dot{\vx} \in \RR^3$ and pullback to the configuration space by the Jacobian pseudoinverse.

\begin{figure}[t]
	\centering
	\includegraphics[width=.5\textwidth]{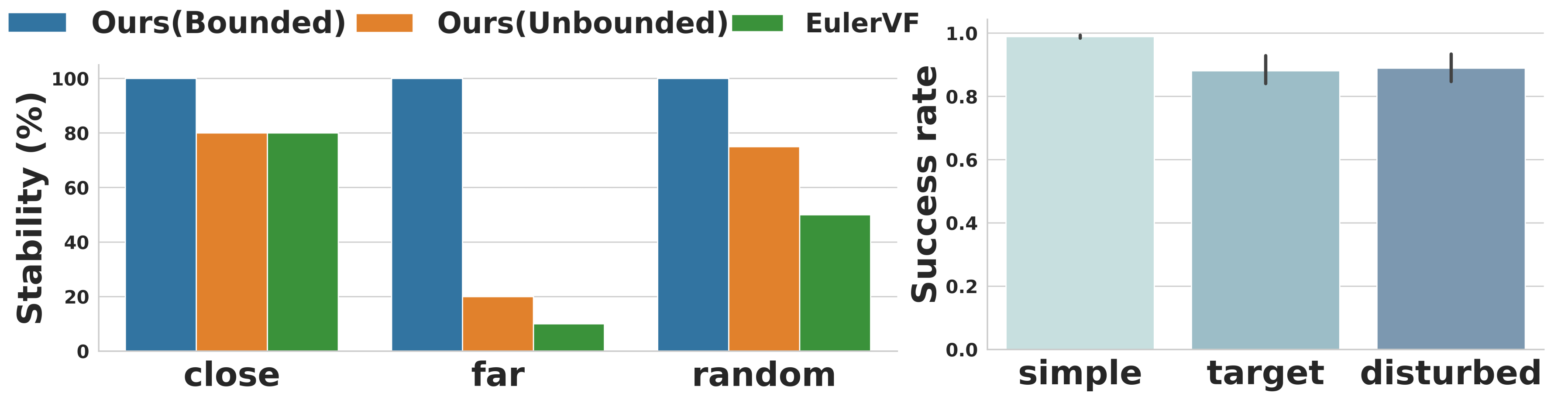}
	\caption{Results for the pouring experiment. Right: simulated experiment results. We compare the stability property of the three models given three possible types of initial configurations (close to the target, far from the target, and random configuration). Left: real robot experiments results.}
	\label{fig:result_pouring}
\end{figure}

We present the results in \cref{fig:peg_hole_results}. We measure the success of the different methods to approach the goal without colliding under different amounts of training data. The vanilla neural network model performed the worst with any amount of trained data. A vanilla-NN is not limiting the family of possible vector fields, thus it may learn vector fields with multiple equilibrium points, limit cycles, or even unstable ones. This results in highly unstable vector fields with poor performance. The results also show the relevance of choosing a good task space representation. Learning in $SE(2)$ outperforms the configuration space approach. The difference in performance might be related to the vector field dimensionality, 5 for the configuration space and 3 for $SE(2)$ and also, with the task itself: as the peg-in-a-hole task is defined in the operational space the $SE(2)$ vector fields fit better the problem. Finally, we observe the benefit of our proposed \gls{inn} w.r.t. vanilla \gls{inn} approach. Given that the vanilla \gls{inn} lacks global stability guarantees, the robot gets stuck in limit cycles and the performance decays.

In conclusion, we have observed that (i) stability guarantees greatly improves the performance of the policy for behavioral cloning problems (ii) representing the vector field in a proper manifold can boost the performance, and (iii) a bounded \gls{inn} guarantees stability, while the unbounded one does not, given $\Phi$ is not diffeomorphic anymore.

\subsection{Learning a pouring task with \texorpdfstring{$SE(3)$}{SE(3)} stable vector fields}
In this experiment, we evaluate the performance of our method on a pouring task (Fig. 1). To properly pour, the robot requires to combine multiple positions and orientation changes. 
First, we compare in simulation our method with Euler angle-based vector fields. We consider two version of our model: One with bounded $\vf_{\vtheta}$, introduced in \cref{sec:invertible_network} and one with a vanilla unbounded \gls{inn} as $\vf_{\vtheta}$~\cite{dinh2016density}. Then, we evaluate the performance of our model in a real robot under target modifications and human disturbances.

For this experiment, we use a 7 DoF Kuka LWR arm. The provided task demonstrations consist of 30 kinesthetic teaching trajectories with a wide variety of initial configurations. We considered different end-effector positions and orientations and trained the three models by behavioral cloning (\cref{alg:train_m_svf}). To control the robot, we apply operational space control~\cite{khatib1987unified} for our proposed model (\cref{fig:architecture}) and position control for the Euler angles vector field. Note that our proposed method adapts to any other type of robot (prismatic joints, parallel robot) by changing the forward kinematics function. We evaluate the three models in three scenarios, robot performance with an initial configuration close to the target, initial configuration far from the target, and random initial configuration. We consider 10 different initial configuration and measure the robot's performance. In the three cases, we measured the stability guarantees of the models (i.e. the guarantee of arriving at the target pose after a certain time). We present the experiment results in \Cref{fig:result_pouring}. From this figure, we can see that our model with the bounded function $f_{\vtheta}$ outperformed the other models in the three cases. These results validate our claims on the requirements of defining a function $f_{\vtheta}$ between the first covers, to guarantee stability in the whole Lie Group. Euler angle-based vector fields perform quite well for the case of close initial configuration. Euler Angles are an undesirable representation for feedback control due to their singularities and non-uniqueness. Nevertheless, we can assume these types of situations are rare close to the target and can perform relatively well. Nevertheless, their performance decay considering initial configurations far from the target. Given the non-uniqueness of the Euler-angles, representing globally stable vector fields in Euler-angles is not possible. In the case of our model with vanilla \gls{inn}, it shows unstable behavior far from the target, while it remains quite stable close to it. Diffeomorphism-based \gls{svf} lack stability guarantees if the function $\Phi$ is not bijective. This lack of bijectiveness is more prone to happen close to the boundaries of the first cover and $\Phi$ remains bijective close to the target, with the guarantee of being stable.

We also evaluate the performance of our model on a real robot, measuring the model's performance under target modifications and human disturbances. To adapt to different target positions, we use the current one $\vx_{\textrm{target}} \in SE(3)$ as the origin of the LogMap ( \cref{fig:architecture}). This allows us to represent the vector fields relative to the current target position. We track the target pot by Optitrack motion capture systems. The control signal is computed in a close-loop at a rate of 100Hz.

For the system evaluation, we predefined 10 different initial configurations covering the whole workspace. The robot holds a glass with 4 balls and we measured the number of balls that enter the pot after executing the trajectory. We considered 3 scenarios: normal execution, physical disturbance, and target modification. 

Looking at the results in~\cref{fig:result_pouring}, it is clear that the robot achieves a very robust performance. In the normal execution, it pours almost all the balls in the pot, given any initial configuration. This result shows the generalization properties of our model: the robot was initialized in a position that does not belong to the demonstration set, but was able to solve the task.
We also tested the system under heavy physical disturbances, including pushing and holding the robot. In this scenario, the performance decays, but the robot was able to succeed most of the time. Finally, we observe the vector field was able to properly adapt to different pot positions. The robot succeeded to put almost all the balls in the pot except for some target positions that were beyond the workspace limits of the robot.

\label{sec:experiments}

\section{Discussion \& Conclusions}

We have proposed a novel Motion Primitive model that can learn stable vector fields on Lie Groups from human demonstrations. Our work extends previous works on modeling stable vector fields to represent them on Lie Groups. The proposed model allows us to generate reactive and stable robot motions for the full pose (orientation and position). Through an extensive evaluation phase, we have validated the modeling decisions to guarantee stability and the importance of representing the vector fields on Lie Groups to properly solve robot tasks.

We have many directions to improve our model. First, the chosen diffeomorphic function $\Phi$ has some limitations. Our proposed model cannot set the sink in the antipodal points, given the map in antipodal points is an identity map. In practice, we can set the attractor in an arbitrary pose by adding a linear transformation that moves the sink. Nevertheless, we consider that this limitation might influence the performance when modeling complex motion skills with significant changes in orientation. In the future, we aim to explore novel functions to represent the diffeomorphism $\Phi$.
The experiments we have carried out focus on the performance evaluation of our proposed stable vector fields. However, these models are of particular interest combined with additional motion skills, such as obstacle avoidance or joint limit avoidance vector fields,  as done in \gls{rmp}~\cite{ratliff2018riemannian} or \gls{cep}~\cite{urain2021composable}. We will investigate how to combine vector fields in future works.

Another possibility is to use the proposed method as a cost function. Indeed, the architecture encodes in itself a Lyapunov-stable potential function. We can use this function as a terminal cost function (value function) or as a cost function in trajectory optimization problems, allowing the integration of additional cost functions. This approach could be beneficial in long-horizon planning problems~\cite{lambert2022learning}.




\section*{ACKNOWLEDGMENT}
This project has received funding from the European Union’s Horizon 2020 research and innovation programmes under grant agreement No. \#820807 (SHAREWORK).
Research presented in this paper has been supported by the German Federal Ministry of Education and Research (BMBF) within a subproject ``Modeling and exploration of the operational area, design of the AI assistance as well as legal aspects of the use of technology'' of the collaborative KIARA project (grant no. 13N16274).


\bibliographystyle{ieeetr}
\bibliography{bibliography}


\end{document}